\documentclass[runningheads]{llncs}
\usepackage[T1]{fontenc}
\usepackage{graphicx}
\usepackage{multirow}
\usepackage{amsmath}
\usepackage{amssymb}
\usepackage{bm}
\usepackage{placeins}
\begin{document}
\title{CETalk: Continuous Valence--Arousal Control for Audio-Driven 3D Talking Head Generation}
\titlerunning{CETalk: Continuous Emotion Control for Talking Heads}
\author{
Peng Jia \and
Li Dai \and
Zhen Xiao \and
Xueliang Liu \and
Jia Li
}
\authorrunning{P. Jia et al.}
\institute{Hefei University of Technology, Hefei 230009, Anhui, China \\
\email{2020214631@mail.hfut.edu.cn}}
\maketitle              % typeset the header of the contribution

\begin{abstract}
Emotional 3D talking head generation aims to synthesize expressive facial animations with accurate lip synchronization. However, existing methods often rely on discrete emotion categories, which fail to capture the continuous evolution of affect. They also overlook the temporal frequency mismatch between audio articulation and emotional expression. In this paper, we propose CETalk, an audio-driven 3D facial animation framework conditioned on continuous Valence--Arousal (VA) representations for fine-grained emotion control. CETalk predicts a sequence of FLAME parameters through three key components: a Dynamic Emotion Modulation Module that adaptively scales emotional intensity using audio-derived cues; a Multi-Scale Temporal Modeling mechanism that employs parallel branches to decouple high-frequency articulatory movements from low-frequency emotional dynamics; and a Dynamic Fusion Mechanism that integrates these multi-scale features via an adaptive gating network. To support training and evaluation, we construct 3D-VA-MEAD, a large-scale dataset with automatically estimated VA annotations and reconstructed 3D facial motions. Extensive experiments demonstrate that CETalk outperforms state-of-the-art methods in both lip-sync accuracy and emotional expressiveness, while enabling smooth and controllable emotion transitions.
\keywords{Talking Head Generation \and 3D Facial Animation \and Continuous Emotion Control}
\end{abstract}

\section{Introduction}

Audio-driven 3D talking head generation aims to synthesize temporally coherent and visually realistic facial animations. With the growing demand for digital humans in applications such as virtual assistants, VR communication, and interactive entertainment~\cite{agarwal2026gaussianheadtalk,cheng2024rita,christoff2023application}, this task has received increasing attention. 

Recent approaches achieve accurate lip synchronization and plausible facial motion by learning mappings from audio features to 3D facial parameters~\cite{cudeiro2019capture,richard2021meshtalk,fan2022faceformer,xing2023codetalker}. Despite these advances, generating emotionally expressive talking heads remains challenging. Existing emotion-aware methods often rely on discrete emotion categories~\cite{peng2023emotalk,prob3d2024}, which limits the ability to represent subtle variations and smooth emotion transitions that naturally occur in human conversations.

\begin{figure*}[t]
  \centering
  \includegraphics[width=\linewidth]{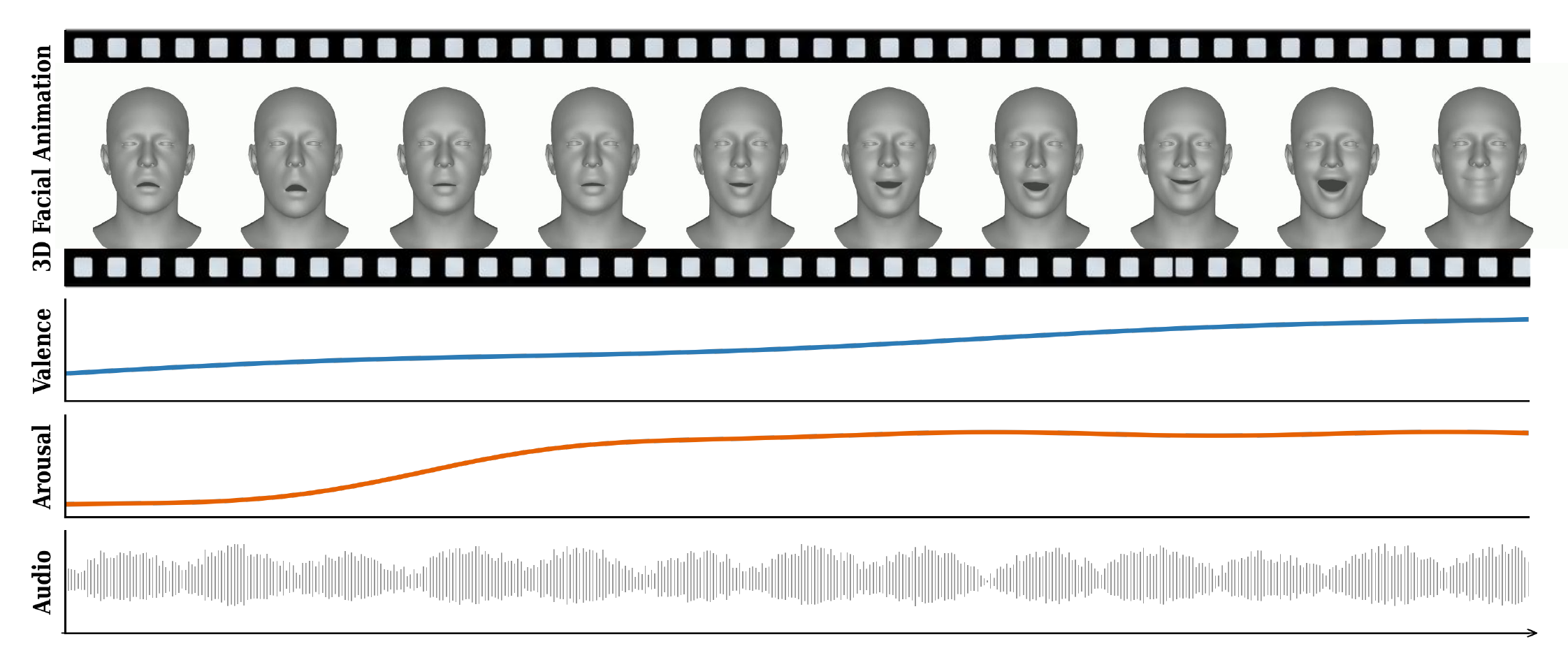}
  \caption{\textbf{Overview of CETalk.} 
Given input audio and continuous Valence--Arousal (VA) labels, the model synthesizes expressive 3D facial animations with accurate lip synchronization and controllable emotional dynamics.}
  \label{fig:teaser}
\end{figure*}

A core challenge lies in modeling the temporal dynamics of emotional expressions during speech. In natural conversations, facial expressions typically evolve gradually over time from onset to peak and decay. Meanwhile, audio articulation and emotional expression operate on markedly different temporal scales: lip movements require high-frequency, frame-level synchronization with the audio, whereas emotional changes evolve more slowly and affect broader facial regions.

To address these challenges, we propose CETalk, an audio-driven 3D talking head framework enabling continuous emotion control via Valence--Arousal conditioning, as illustrated in Fig.~\ref{fig:teaser}. It comprises three key components: (1) the Dynamic Emotion Modulation Module (DEMM) that adaptively scales emotional intensity from speech cues, (2) the Multi-Scale Temporal Modeling (MSTM) mechanism with parallel high-frequency (audio articulation) and low-frequency (emotional dynamics) branches to decouple their temporal properties, and (3) the Dynamic Fusion Mechanism (DFM) that adaptively integrates multi-scale features for coherent and expressive animations. Furthermore, we construct 3D-VA-MEAD, a large-scale dataset derived from MEAD~\cite{wang2020mead}, providing reconstructed 3D facial motion and frame-level VA annotations.

Our contributions are summarized as follows:
\begin{enumerate}
    \item We propose CETalk, an audio-driven 3D talking head generation framework that enables continuous VA-based emotion control for more expressive and controllable facial animation.
    
    \item We introduce a Multi-Scale Temporal Modeling (MSTM) mechanism that explicitly disentangles and models the heterogeneous temporal dynamics of audio articulation and emotional expression.
    
    \item We construct 3D-VA-MEAD, a large-scale dataset with reconstructed 3D facial motion and frame-level VA annotations to support research on emotion-aware talking head generation.
\end{enumerate}

\section{Related Work}

\begin{figure}[t]
\centering
\includegraphics[width=0.8\linewidth]{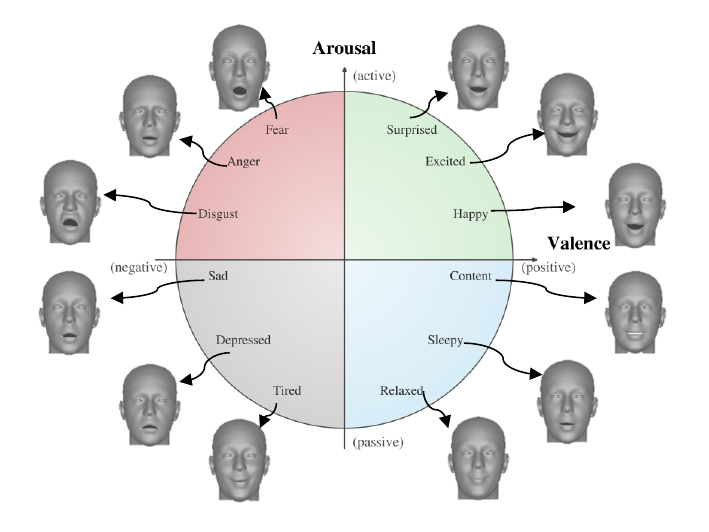}
\caption{\textbf{Valence--Arousal affective space for facial expressions.}
Emotions are represented in a continuous two-dimensional space defined by valence (positive–negative) and arousal (active–passive). Different regions correspond to distinct emotional states (e.g., happy, surprised, sad), illustrating how facial expressions vary smoothly across the VA plane.}
\label{fig:va_faces}
\end{figure}

\subsection{3D Talking Head Generation}

Audio-driven 3D talking head generation aims to synthesize realistic facial motion. Early methods typically employed neural mappings that convert audio features into facial geometry, such as VOCA~\cite{cudeiro2019capture} for predicting 3D vertex positions and MeshTalk~\cite{richard2021meshtalk}, which models facial dynamics by disentangling audio-related and audio-independent motions. More recent approaches adopt sequence modeling to capture long-range dependencies: FaceFormer~\cite{fan2022faceformer} utilizes an autoregressive Transformer, CodeTalker~\cite{xing2023codetalker} incorporates discrete motion priors, and diffusion-based models such as FaceDiffuser~\cite{stan2023facediffuser}, DiffPoseTalk~\cite{chu2024diffposetalk}, and RealTalk~\cite{ji2024realtalk} further improve motion diversity and realism. Despite these improvements, most methods primarily focus on facial motion while largely neglecting continuous emotional dynamics. Some attempts to incorporate emotion rely on discrete emotion labels for style control~\cite{peng2023emotalk,prob3d2024},  which fail to capture the gradual evolution of affect in natural speech, resulting in limited expressiveness.

\subsection{Facial Expression Representations}

Facial expressions are typically represented using three main paradigms: discrete emotion categories, facial action units, and continuous affective models. Discrete emotion models based on Ekman's basic emotions~\cite{ekman1993facial} are intuitive but struggle to capture subtle affective variations. The Facial Action Coding System (FACS)~\cite{ekman1978facial} provides fine-grained modeling by decomposing facial movements into action units corresponding to muscle activations, but it requires complex parameter specification. In contrast, continuous affective models, particularly the Valence--Arousal (VA) framework~\cite{russell1980circumplex}, represent emotions as points in a two-dimensional continuous space, where valence indicates emotional positivity or negativity and arousal reflects the level of emotional activation (see Fig.~\ref{fig:va_faces}). Such a representation naturally supports smooth transitions between emotional states and is therefore well suited for modeling the gradually evolving emotional dynamics in audio-driven facial animation.

\section{Method}

\subsection{Overview}
\label{sec:overview}

Given an input audio sequence, the objective of 3D talking head generation is to synthesize a temporally coherent sequence of facial motions. Let $\mathbf{A}=\{a_t\}_{t=1}^{T}$ denote the input audio sequence, where $T$ is the number of video frames. In addition, we are given a speaker identity embedding $\mathbf{S}$ and an emotion condition $\mathbf{E}_{\text{cond}} \in \mathbb{R}^{T \times 2}$ defined in the valence--arousal space. The goal is to predict a sequence of facial motion parameters $\mathbf{P} \in \mathbb{R}^{T \times D_p}$, where $D_p$ denotes the dimensionality of the FLAME~\cite{li2017learning} parameter vector for each frame.

Formally, the task can be formulated as learning a mapping
\begin{equation}
\mathbf{P} = \mathcal{F}(\mathbf{A}, \mathbf{S}, \mathbf{E}_{\text{cond}}),
\end{equation}
where $\mathcal{F}(\cdot)$ denotes the proposed model.

\begin{figure*}[t]
  \centering
  \includegraphics[width=\linewidth]{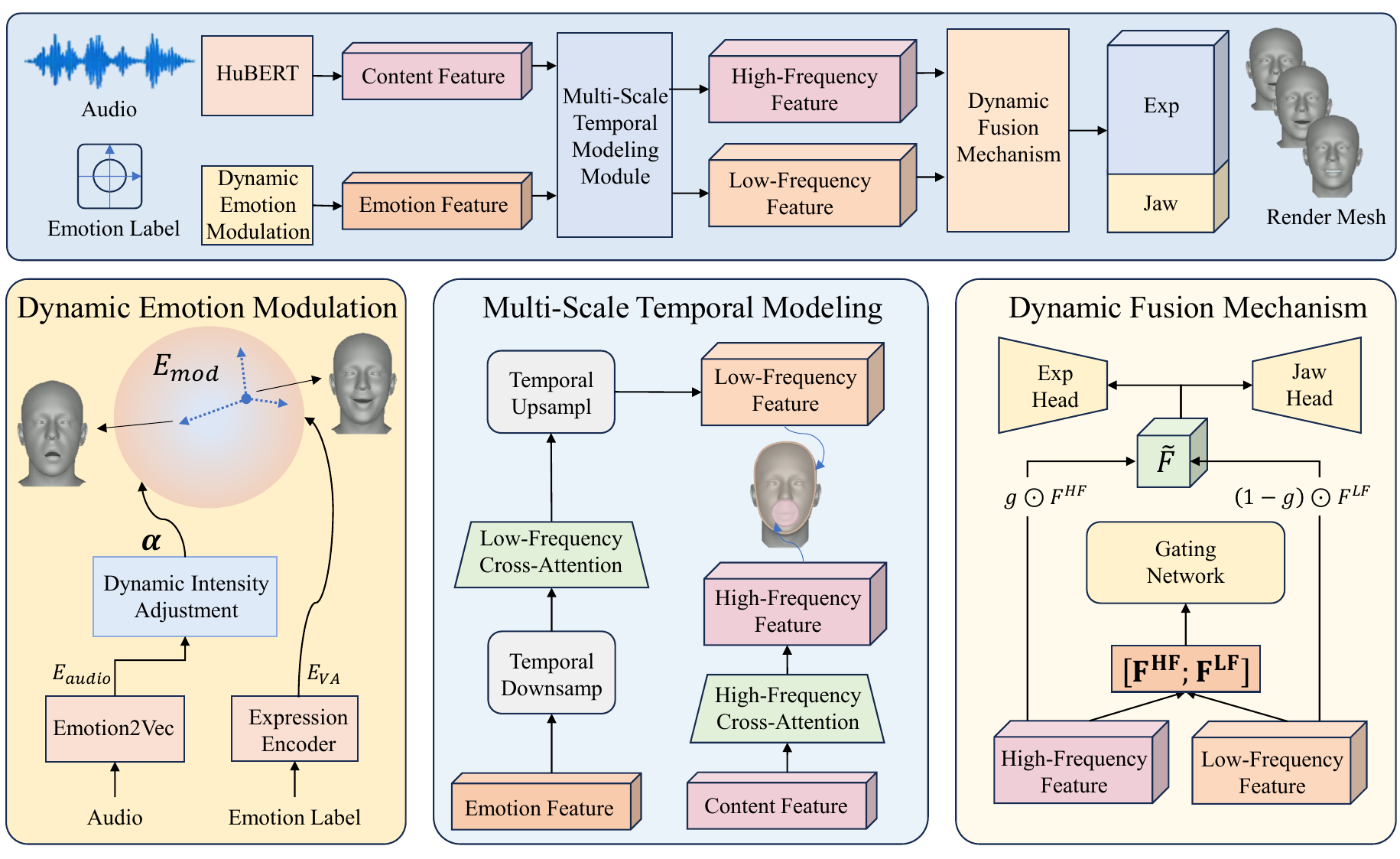}
  \caption{\textbf{Overview of the CETalk framework.} 
Given input audio $\mathbf{A}$, speaker identity $\mathbf{S}$, and continuous emotion condition $\mathbf{E}_{cond}$ in the Valence--Arousal space, the model predicts a sequence of FLAME parameters $\mathbf{P}$. 
The Dynamic Emotion Modulation module transforms $\mathbf{E}_{cond}$ into a dynamic emotion embedding $\mathbf{E}_{mod}$, the Multi-Scale Temporal Modeling module extracts high-frequency articulatory features $\mathbf{F}^{HF}$ and low-frequency emotional features $\mathbf{F}^{LF}$, and the Dynamic Fusion Mechanism integrates them to generate expressive 3D facial animations.}
  \label{fig:framework}
\end{figure*}

As illustrated in Fig.~\ref{fig:framework}, the proposed CETalk framework consists of three key components. First, the Dynamic Emotion Modulation Module (Sec.~\ref{sec:dem}) extracts affective cues from the input audio and modulates the static emotional condition $\mathbf{E}_{\text{cond}}$ into a dynamic representation $\mathbf{E}_{\text{mod}}$. Second, the Multi-Scale Temporal Modeling Module (Sec.~\ref{sec:mtm}) processes the audio features through parallel temporal branches to disentangle high-frequency articulatory dynamics from low-frequency emotional variations. Finally, the Dynamic Fusion Mechanism (Sec.~\ref{sec:dfm}) adaptively integrates these multi-scale representations to generate the final sequence of facial motion parameters $\mathbf{P}$.

\subsection{Dataset Construction}
\label{sec:dataset}

Due to the lack of publicly available 3D talking head datasets annotated with VA labels, we construct 3D-VA-MEAD, a large-scale emotional 3D facial motion dataset based on the MEAD~\cite{wang2020mead} dataset.

Specifically, we first employ a pretrained emotion recognition model~\cite{toisoul2021estimation} to estimate frame-level valence and arousal coefficients from the input videos. This produces an initial emotion sequence
$\mathbf{E}_{\text{raw}} = \{\mathbf{e}^{\text{raw}}_t\}_{t=1}^{T}$,
where $\mathbf{e}^{\text{raw}}_t \in \mathbb{R}^{2}$ denotes the valence--arousal vector of frame $t$. 
Meanwhile, we recover the corresponding 3D facial motion by applying a monocular 3D face reconstruction method~\cite{danecek2022emoca}, which estimates the FLAME parameter sequence from the source videos.

To mitigate temporal instability in the regressed predictions, we apply a one-dimensional moving average filter for temporal smoothing. Formally, the refined emotion vector $\mathbf{e}_t$ at timestep $t$ is computed as:
\begin{equation}
\mathbf{e}_t = \frac{1}{K} \sum_{i = -\lfloor K/2 \rfloor}^{\lfloor K/2 \rfloor} \mathbf{e}^{raw}_{t+i},
\end{equation}
where $K$ represents the temporal window size. The final continuous emotion condition used for model training is then defined as the sequence of these smoothed vectors: $\mathbf{E}_{cond} = \{\mathbf{e}_t\}_{t=1}^T \in \mathbb{R}^{T \times 2}$.

\subsection{Dynamic Emotion Modulation Module}
\label{sec:dem}

To bridge the gap between static emotional specifications and the inherently dynamic nature of facial expressions, we introduce a dynamic emotion modulation module that transforms Valence--Arousal conditions into temporally evolving affective representations. As illustrated in Fig.~\ref{fig:framework}, this module consists of three components: an expression encoder $E_{\text{exp}}(\cdot)$, an audio--emotion encoder $E_{\text{emotion}}(\cdot)$ based on emotion2vec~\cite{ma2023emotion2vec}, and a dynamic intensity adjustment module $\mathcal{M}(\cdot)$.

Given the emotion condition $\mathbf{E}_{\text{cond}} \in \mathbb{R}^{T \times 2}$, we first encode the VA signals using the expression encoder to obtain a base emotional representation $\mathbf{E}_{\text{VA}} = E_{\text{exp}}(\mathbf{E}_{\text{cond}}) \in \mathbb{R}^{T \times D}$, where $D$ denotes the embedding dimensionality. Meanwhile, frame-level affective cues are extracted from the input audio $\mathbf{A}$ using $E_{\text{emotion}}$, yielding $\mathbf{E}_{\text{audio}} = E_{\text{emotion}}(\mathbf{A}) \in \mathbb{R}^{T \times D}$. Based on these audio features, the adjustment module predicts a sequence of temporal scaling coefficients $\boldsymbol{\alpha} = \mathcal{M}(\mathbf{E}_{\text{audio}}) \in \mathbb{R}^{T \times 1}$ to capture audio-driven emotional intensity.

The final dynamically modulated emotion embedding is obtained through element-wise scaling:
\begin{equation}
\mathbf{E}_{\text{mod}} = \boldsymbol{\alpha} \odot \mathbf{E}_{\text{VA}},
\end{equation}
where $\odot$ denotes element-wise multiplication.

\subsection{Multi-Scale Temporal Modeling Module}
\label{sec:mtm}

Audio-driven facial animation exhibits an inherent temporal frequency mismatch: articulatory mouth movements require frame-level alignment with audio, whereas emotional expressions evolve more slowly over longer temporal spans. To decouple these dynamics, we introduce a Multi-Scale Temporal Modeling Module with two parallel branches that capture high-frequency (HF) articulatory motions and low-frequency (LF) expression dynamics.

Given an input audio sequence $\mathbf{A}$, we first extract content features $\mathbf{H}^{a} = E_{\text{content}}(\mathbf{A}) \in \mathbb{R}^{T \times D}$ using a HuBERT encoder~\cite{baevski2020wav2vec}. The HF branch focuses on modeling fine-grained articulatory motion that must remain tightly synchronized with audio. We compute the high-frequency motion representation using cross-attention between the speaker embedding $\mathbf{S}$ and the audio features $\mathbf{H}^{a}$. The resulting high-frequency motion representation $\mathbf{F}^{HF} \in \mathbb{R}^{T \times D}$ is computed as
\begin{equation}
\mathbf{F}^{HF} =
\text{Softmax}\!\left(\frac{\mathbf{Q}^{HF}(\mathbf{K}^{HF})^{T}}{\sqrt{d}}\right)\mathbf{V}^{HF},
\end{equation}
where $\mathbf{Q}^{HF}=\mathbf{S}\mathbf{W}_q^{hf}$, $\mathbf{K}^{HF}=\mathbf{H}^{a}\mathbf{W}_k^{hf}$, and $\mathbf{V}^{HF}=\mathbf{H}^{a}\mathbf{W}_v^{hf}$.

In contrast, the LF branch models slowly varying emotional dynamics. We first apply a temporal downsampling operator $\mathcal{D}(\cdot)$ with stride $k$ to obtain compressed representations $\mathbf{H}^{a}_{LF}=\mathcal{D}(\mathbf{H}^{a})$ and $\mathbf{E}_{LF}=\mathcal{D}(\mathbf{E}_{\text{mod}})$. Cross-attention is then performed at this coarse temporal scale using the concatenated emotional context $[\mathbf{S};\mathbf{E}_{LF}]$ as the query and $\mathbf{H}^{a}_{LF}$ as the key-value features. The resulting representation is finally restored to the original temporal resolution through an upsampling operator $\mathcal{U}(\cdot)$, yielding the low-frequency feature sequence $\mathbf{F}^{LF}\in\mathbb{R}^{T\times D}$.

Through this explicit temporal scale separation, the LF branch captures dynamic emotional variations, while the HF branch preserves precise audio-driven articulatory motions, enabling more natural and temporally coherent facial animation generation.

\subsection{Dynamic Fusion Mechanism}
\label{sec:dfm}

To effectively integrate the complementary representations learned by the high-frequency and low-frequency branches, we introduce a dynamic fusion mechanism that adaptively balances information across temporal scales. Given the features $\mathbf{F}^{HF}, \mathbf{F}^{LF} \in \mathbb{R}^{T \times D}$, we first derive channel-wise adaptive gating weights $\mathbf{g} \in \mathbb{R}^{T \times D}$ via a lightweight gating network:
\begin{equation}
\mathbf{g} = \sigma \left( \text{MLP}_{gate}( [\mathbf{F}^{HF}; \mathbf{F}^{LF}] ) \right),
\end{equation}
where $[ \cdot ; \cdot ]$ denotes concatenation along the channel dimension and $\sigma$ is the Sigmoid activation. The final unified temporal representation $\tilde{\mathbf{F}} \in \mathbb{R}^{T \times D}$ is then computed through an element-wise weighted combination:
\begin{equation}
\tilde{\mathbf{F}} = \mathbf{g} \odot \mathbf{F}^{HF} + (\mathbf{1} - \mathbf{g}) \odot \mathbf{F}^{LF}.
\end{equation}

Finally, $\tilde{\mathbf{F}}$ is fed into parallel linear projection heads to estimate the expression coefficients $\boldsymbol{\Psi} \in \mathbb{R}^{T \times 50}$ and jaw coefficients $\boldsymbol{\Theta} \in \mathbb{R}^{T \times 3}$. The complete sequence of 3D facial motion parameters $\mathbf{P} \in \mathbb{R}^{T \times 53}$ is obtained by concatenating these predictions: $\mathbf{P} = [ \boldsymbol{\Psi}; \boldsymbol{\Theta} ]$.

\subsection{Loss Function}
\label{sec:loss}

We train the model using a combination of parameter, vertex, and temporal smoothness losses to ensure accurate facial motion reconstruction and stable animation.

\textbf{Parameter loss.} Let $\hat{\boldsymbol{\Psi}}$ and $\hat{\boldsymbol{\Theta}}$ denote the predicted expression and jaw parameter sequences, respectively. We minimize the L1 reconstruction error against the ground truth $\boldsymbol{\Psi}$ and $\boldsymbol{\Theta}$:
\begin{equation}
\mathcal{L}_{param} = \|\hat{\boldsymbol{\Psi}} - \boldsymbol{\Psi}\|_1 + \|\hat{\boldsymbol{\Theta}} - \boldsymbol{\Theta}\|_1.
\end{equation}

\textbf{Vertex loss.}
To enforce geometric consistency, we compute vertex reconstruction loss. Let $\hat{\mathbf{V}}, \mathbf{V} \in \mathbb{R}^{T \times N \times 3}$ denote the predicted and ground-truth vertex sequences obtained from the FLAME model, where $T$ is the number of frames and $N$ is the number of vertices. The loss is defined as
\begin{equation}
\mathcal{L}_{vert}
=
\frac{1}{TN}
\|
\hat{\mathbf{V}} - \mathbf{V}
\|_1 .
\end{equation}

\textbf{Velocity loss.}
To encourage temporally smooth motion, we penalize the difference between predicted and ground-truth vertex velocities. Let
$\Delta \hat{\mathbf{V}} = \hat{\mathbf{V}}_{2:T} - \hat{\mathbf{V}}_{1:T-1}$ and
$\Delta \mathbf{V} = \mathbf{V}_{2:T} - \mathbf{V}_{1:T-1}$. The velocity loss is
\begin{equation}
\mathcal{L}_{vel}
=
\frac{1}{(T-1)N}
\|
\Delta \hat{\mathbf{V}} - \Delta \mathbf{V}
\|_F^2 .
\end{equation}

The overall objective is the weighted sum of all terms:
\begin{equation}
\mathcal{L}
=
\lambda_{param} \mathcal{L}_{param}
+
\lambda_{vert} \mathcal{L}_{vert}
+
\lambda_{vel} \mathcal{L}_{vel}.
\end{equation}

\section{Experiments}
\subsection{Experimental Settings}

\textbf{Datasets.} 
We evaluate our model on three datasets to verify its performance and generalization ability.
MEAD~\cite{wang2020mead} is a large-scale emotional dataset containing 60 actors performing eight emotions at three intensity levels. It serves as our primary dataset for constructing 3D-VA-MEAD (Sec.~\ref{sec:dataset}) and conducting the main evaluations.
RAVDESS~\cite{livingstone2018ravdess} consists of 24 professional actors expressing eight emotions through speech and song.
HDTF~\cite{zhang2021flow} comprises approximately 16 hours of high-resolution in-the-wild videos. For all datasets, the FLAME parameters are extracted using EMOCA~\cite{danecek2022emoca}.

\textbf{Implementation Details.} 
We implement CETalk in PyTorch and train the model on an NVIDIA RTX 4090 GPU. The model is trained using the AdamW optimizer with an initial learning rate of $1\times10^{-4}$. For the objective function, we empirically set the loss weights as $\lambda_{\mathrm{param}}=1.0$, $\lambda_{\mathrm{vert}}=1.0$, and $\lambda_{\mathrm{vel}}=0.5$.

\textbf{Baselines.} 
We compare CETalk against several state-of-the-art audio-driven 3D talking head generation methods, including FaceFormer~\cite{fan2022faceformer}, CodeTalker~\cite{xing2023codetalker}, EMOTE~\cite{emote2023}, ProbTalk3D~\cite{prob3d2024}, UniTalker~\cite{unitalker2024}, and DEEPTalk~\cite{deeptalk2025}.

\subsection{Quantitative Evaluation}
\label{sec:quant_eval}

\textbf{Evaluation Metrics.}
Following standard evaluation protocols in 3D facial animation~\cite{fan2022faceformer,xing2023codetalker}, we adopt three geometric metrics, including Lip Vertex Error (LVE), Mean Vertex Error (MVE), and Emotional Vertex Error (EVE). Specifically, LVE measures the accuracy of lip motion, MVE evaluates the overall vertex reconstruction error across the full face, and EVE focuses on expression-related regions to assess the quality of emotional facial dynamics.

To further evaluate continuous emotion control, we additionally introduce two affective alignment metrics for both valence and arousal: Root Mean Square Error (RMSE)~\cite{danecek2022emoca,mollahosseini2017affectnet} and Sign Agreement (SAGR)~\cite{danecek2022emoca,kollias2020deep}. RMSE measures the numerical deviation between the predicted and ground-truth affective signals, while SAGR evaluates whether they share the same emotional polarity at each frame.

Let $\hat{y}_t$ and $y_t$ denote the predicted and ground-truth emotional values (valence or arousal) at frame $t$, respectively, over a sequence of $T$ frames. RMSE is defined as
\begin{equation}
\text{RMSE} = \sqrt{\frac{1}{T} \sum_{t=1}^{T} (\hat{y}_t - y_t)^2}.
\label{eq:rmse}
\end{equation}

The SAGR measures whether the predicted and ground-truth emotional values share the same sign at each frame:
\begin{equation}
\text{SAGR} = \frac{1}{T} \sum_{t=1}^{T} \delta \bigl( \operatorname{sign}(\hat{y}_t), \operatorname{sign}(y_t) \bigr),
\label{eq:sagr}
\end{equation}
where $\operatorname{sign}(x) = 1$ if $x > 0$, $-1$ if $x < 0$, and $0$ if $x = 0$, and $\delta(a, b) = 1$ if $a = b$ else $0$. Higher SAGR indicates better consistency in emotional polarity.

To obtain the affective signals for evaluation, we first render the predicted 3D facial animations into 2D images using GAGAvatar~\cite{chu2024gagavatar}. We then apply a pretrained affect estimation model~\cite{toisoul2021estimation} to extract frame-wise valence and arousal values for metric computation.

% --- Table 1: Geometric Metrics ---
\begin{table*}[t]
\centering
\caption{Quantitative comparison of facial motion accuracy ($mm$). Lower values ($\downarrow$) indicate better performance. Best results are in \textbf{bold} and second-best are \underline{underlined}.}
\label{tab:quantitative}
\resizebox{\textwidth}{!}{
\begin{tabular}{lccc|ccc|cc}
\hline
\multirow{2}{*}{Method} 
& \multicolumn{3}{c|}{MEAD} 
& \multicolumn{3}{c|}{RAVDESS} 
& \multicolumn{2}{c}{HDTF} \\
\cline{2-9}
& LVE$\downarrow$ & MVE$\downarrow$ & EVE$\downarrow$
& LVE$\downarrow$ & MVE$\downarrow$ & EVE$\downarrow$
& LVE$\downarrow$ & MVE$\downarrow$ \\
\hline
FaceFormer~\cite{fan2022faceformer} & 19.1505 & 17.2842 & 2.5844 & 16.0241 & 13.7128 & 2.3885 & 10.3196 & 11.6200 \\
CodeTalker~\cite{xing2023codetalker} & 19.3656 & 17.0043 & 2.5571 & 15.9058 & 13.7034 & 2.5207 & 11.0166 & 11.3826 \\
EMOTE~\cite{emote2023}      & 10.2828 & 13.4281 & 3.4535 & 7.9951 & 13.4142 & 5.9953 & \underline{5.2671} & 11.9896 \\
ProbTalk3D~\cite{prob3d2024} & 11.5005 & 12.7797 & 2.2024 & \textbf{7.7616} & 12.1496 & 3.5825 & 5.6203 & \underline{10.9151} \\
UniTalker~\cite{unitalker2024}  & \underline{9.5945} & 12.2777 & 2.1202 & 10.1716 & 11.6141 & 1.8787 & 7.2121 & 10.6160 \\
DEEPTalk~\cite{deeptalk2025}   & 12.3237 & \underline{11.5817} & \underline{1.5566} & 10.6610 & \textbf{10.5335} & \underline{1.6201} & 11.8344 & 16.4693 \\
\hline
\textbf{Ours} & \textbf{7.4772} & \textbf{9.9059} & \textbf{1.3256}
              & \underline{7.8145} & \underline{10.7293} & \textbf{1.5475}
              & \textbf{4.7408} & \textbf{8.9010} \\
\hline
\end{tabular}
}
\end{table*}

% --- Table 2: Emotion Metrics ---
\begin{table}[t]
\centering
\caption{Quantitative evaluation of emotion controllability using RMSE and SAGR.}
\label{tab:emotion_control}
\begin{tabular}{l|cc|cc}
\hline
\multirow{2}{*}{Method} & \multicolumn{2}{c|}{RMSE $\downarrow$} & \multicolumn{2}{c}{SAGR $\uparrow$} \\
\cline{2-5}
& Valence & Arousal & Valence & Arousal \\
\hline
EMOTE~\cite{emote2023}       & 0.2133 & 0.1219 & 0.7909 & 0.9693 \\
ProbTalk3D~\cite{prob3d2024}  & \underline{0.1826} & \underline{0.1059} & \underline{0.8491} & 0.9861 \\
UniTalker~\cite{unitalker2024}   & 0.3118 & 0.1537 & 0.6589 & \underline{0.9870} \\
DEEPTalk~\cite{deeptalk2025}    & 0.3156 & 0.1310 & 0.7720 & 0.9833 \\
\hline
\textbf{Ours} & \textbf{0.1223} & \textbf{0.0747} & \textbf{0.9247} & \textbf{0.9906} \\
\hline
\end{tabular}
\end{table}

\textbf{Facial Motion Accuracy.}
Table~\ref{tab:quantitative} reports the quantitative results of facial motion prediction. On MEAD, CETalk achieves the lowest errors with 7.4772 LVE and 9.9059 MVE, outperforming all baselines, including the classical method FaceFormer~\cite{fan2022faceformer} and recent approaches such as ProbTalk3D~\cite{prob3d2024}, UniTalker~\cite{unitalker2024}, and DEEPTalk~\cite{deeptalk2025}. 
On RAVDESS, CETalk maintains competitive performance with 7.8145 LVE and 10.7293 MVE. 
On the in-the-wild HDTF dataset, CETalk further achieves the lowest errors with 4.7408 LVE and 8.9010 MVE, demonstrating strong cross-dataset generalization. 

\textbf{Emotion Controllability.}
Table~\ref{tab:emotion_control} reports the quantitative results for continuous emotion control. CETalk achieves the lowest RMSE for both valence and arousal, indicating that the generated facial motions follow the target affective trajectories more accurately than competing methods. In addition, CETalk obtains the highest SAGR scores on both dimensions, showing superior consistency in emotional polarity. These results confirm that CETalk not only produces accurate facial motions, but also enables precise and stable continuous emotion modulation.

\begin{figure*}[t!]
  \centering
  \includegraphics[height=0.7\textheight, width=\textwidth, keepaspectratio]{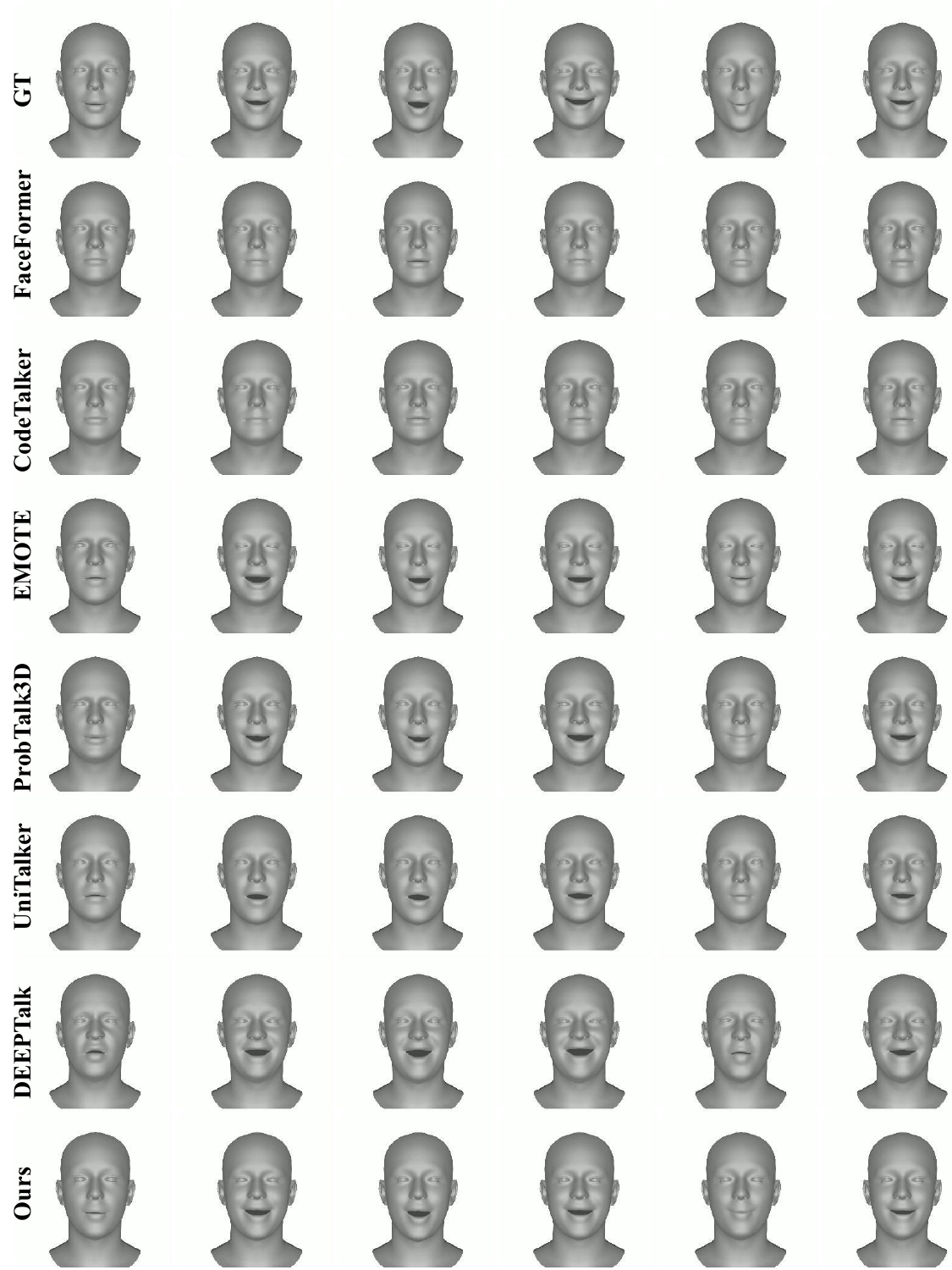}
  \caption{Qualitative comparison of generated 3D talking head animations.}
  \label{fig:qualitative}
\end{figure*}

\begin{figure*}[t!]
  \centering
  \includegraphics[height=0.6\textheight, width=\textwidth, keepaspectratio]{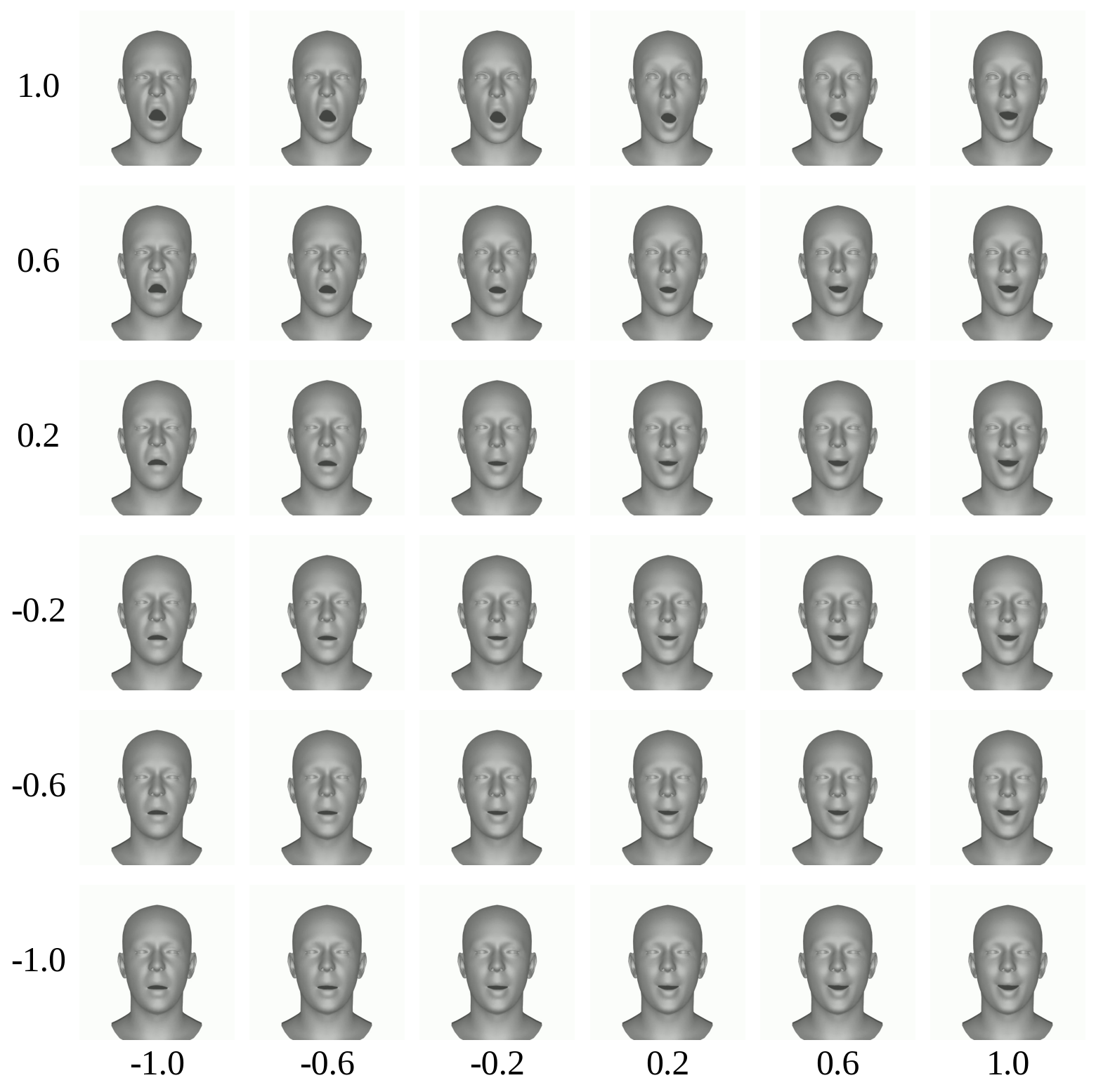}
  \caption{\textbf{Continuous emotion controllability via VA-space interpolation.}
With the same audio clip, we linearly interpolate the input Valence--Arousal condition from $(-1,-1)$ to $(1,1)$.}
  \label{fig:interp}
\end{figure*}

\begin{figure*}[t]
\centering
\includegraphics[width=\linewidth]{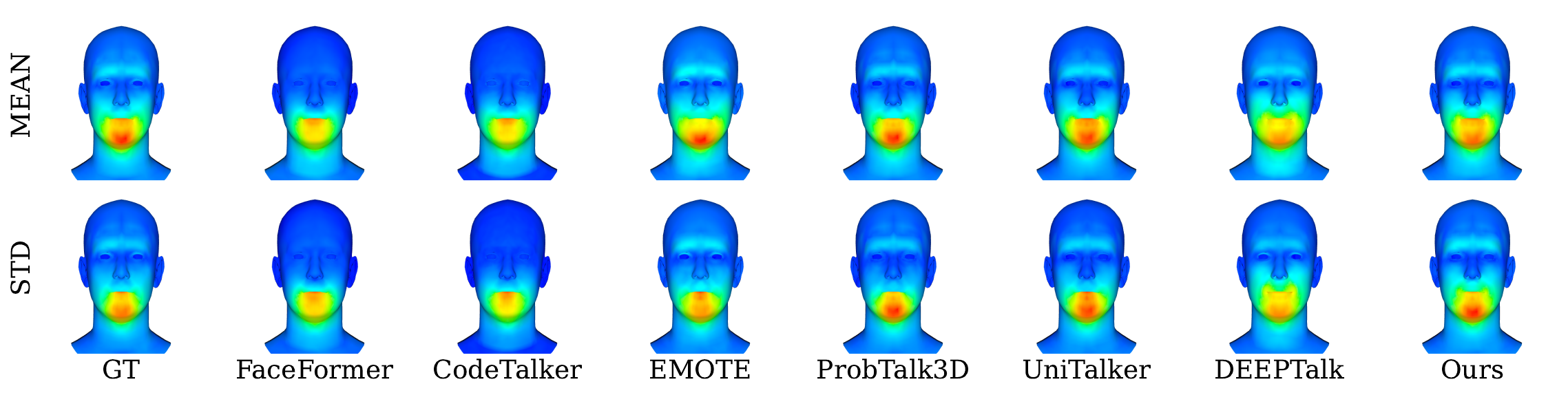}
\caption{\textbf{Heatmap comparison of the mean and standard deviation}: Ground Truth versus animations generated by different methods.}
\label{fig:heatmap}
\end{figure*}

\subsection{Qualitative Evaluation}

We qualitatively compare CETalk with prior methods in Fig.~\ref{fig:qualitative}. CETalk achieves accurate lip synchronization and generates natural, expressive facial motions. Compared with FaceFormer~\cite{fan2022faceformer} and CodeTalker~\cite{xing2023codetalker}, our method produces richer and more coherent expressive dynamics. It also yields smoother expression transitions than DEEPTalk~\cite{deeptalk2025}, leading to more natural animations. These results validate the effectiveness of the proposed multi-scale architecture in maintaining both temporal stability and expressive diversity.

We further demonstrate continuous emotion control through interpolation in the Valence--Arousal space (Fig.~\ref{fig:interp}). As the input VA condition varies smoothly, the generated expressions transition gradually in intensity while preserving lip-sync accuracy. This confirms that CETalk supports fine-grained and temporally coherent emotion control while preserving affective semantics.

Fig.~\ref{fig:heatmap} presents the mean and standard deviation heatmaps of vertex displacements. CETalk closely matches the Ground Truth in mean motion magnitude and exhibits more realistic temporal variation in the STD heatmaps. In contrast, FaceFormer~\cite{fan2022faceformer} and CodeTalker~\cite{xing2023codetalker} show clear over-smoothing, leading to rigid and less expressive animations. These results demonstrate that CETalk better reconstructs natural motion dynamics while preserving both lip-sync accuracy and emotional expressiveness.

\subsection{Ablation Study}

We conduct ablation studies on MEAD to evaluate the contribution of each core component. As shown in Table~\ref{tab:ablation}, the full model achieves the best results on all metrics. Removing DEMM causes the largest degradation, with EVE increasing from 1.3256 to 2.4367, indicating that dynamic intensity modulation is essential for modeling temporal emotional variation. Removing MSTM increases both LVE and EVE, confirming the benefit of explicitly decoupling articulatory and emotional dynamics. Replacing DFM with simple addition also degrades performance, particularly in LVE and MVE, showing that adaptive fusion is important for balancing fast audio-driven motion and slow emotional changes.

% --- Table: Ablation Study ---
\begin{table}[h]
\centering
\caption{Ablation study on the MEAD dataset. LVE, MVE, and EVE are reported to evaluate the contribution of each component.}
\label{tab:ablation}
\begin{tabular}{l|ccc}
\hline
Method & LVE $\downarrow$ & MVE $\downarrow$ & EVE $\downarrow$ \\
\hline
Ours & \textbf{7.4772} & \textbf{9.9059} & \textbf{1.3256} \\
w/o DEMM          & 9.6963          & 12.3722         & 2.4367          \\
w/o MSTM           & 8.0398          & 10.3324         & 1.4123          \\
w/o DFM           & 8.1931          & 10.6743         & 1.6406          \\
\hline
\end{tabular}
\end{table}

\section{Conclusion}

We proposed CETalk, an audio-driven 3D talking head generation framework with continuous Valence--Arousal control. By modeling emotion in a continuous affective space, CETalk enables fine-grained and temporally coherent facial animation beyond discrete emotion categories. CETalk combines dynamic emotion modulation, multi-scale temporal modeling, and adaptive feature fusion to jointly capture audio-driven articulation and emotional dynamics. We also construct 3D-VA-MEAD, a large-scale dataset with reconstructed 3D facial motion and continuous VA annotations. Extensive experiments demonstrate that CETalk achieves strong performance in lip synchronization, facial motion accuracy, and emotion controllability, validating the effectiveness of continuous affective modeling for expressive 3D talking head generation.

\bibliographystyle{splncs04}
\bibliography{mybib}   % mybib 是你的 .bib 文件名，不要写 .bib 后缀

\end{document}